\documentclass[acmsmall]{acmart}
\usepackage{enumerate}
\usepackage{dsfont}

\AtBeginDocument{%
  \providecommand\BibTeX{{%
    \normalfont B\kern-0.5em{\scshape i\kern-0.25em b}\kern-0.8em\TeX}}}

\newcommand{\revision}[1]{{#1}}

\begin{document}

\title{
How to Train your Quadrotor: A Framework for Consistently Smooth and Responsive Flight Control via Reinforcement Learning
}

\author{Siddharth Mysore}
\authornote{Both authors contributed equally to this research.}
\email{sidmys@bu.edu}
\author{Bassel Mabsout}
\authornotemark[1]
\email{bmabsout@bu.edu}
\affiliation{%
  \institution{Boston University}
  \city{Boston}
  \state{Massachusetts}
  \postcode{02215}
}

\author{Kate Saenko}
\affiliation{%
  \institution{Boston University \& MIT-IBM Watson AI Lab}
  \city{Boston}
  \state{Massachusetts}
  \postcode{02215}
}
\email{saenko@bu.edu}

\author{Renato Mancuso}
\affiliation{%
  \institution{Boston University}
  \city{Boston}
  \state{Massachusetts}
  \postcode{02215}
}
\email{rmancuso@bu.edu}







\begin{abstract}
    We focus on the problem of reliably training Reinforcement Learning (RL) models (agents) for stable low-level control in embedded systems and test our methods on a high-performance, custom-built quadrotor platform.
    A common but often under-studied problem in developing RL agents for continuous control is that the control policies developed are not always smooth.
    This lack of smoothness can be a major problem when learning controllers  
    as it can result in control instability and hardware failure.
    
    Issues of noisy control are further accentuated when training RL agents in simulation due to simulators ultimately being imperfect representations of reality --- what is known as the \emph{reality gap}.
    To combat issues of instability in RL agents, we propose a systematic framework, `REinforcement-based transferable Agents through Learning' (RE+AL), for designing simulated training environments which preserve the quality of trained agents when transferred to real platforms.
    {RE+AL is an evolution of the Neuroflight infrastructure detailed in technical reports prepared by members of our research group.
    Neuroflight is a state-of-the-art framework for training RL agents for low-level attitude control. 
    RE+AL improves and completes Neuroflight by solving a number of important limitations that hindered the deployment of Neuroflight to real hardware. We benchmark RE+AL on the NF1 racing quadrotor developed as part of Neuroflight}.
    We demonstrate that RE+AL significantly mitigates the previously observed issues of smoothness in RL agents. Additionally, RE+AL is shown to consistently train agents that are flight-capable and with minimal degradation in controller quality upon transfer. 
    {
    RE+AL agents also learn to perform better than a tuned PID controller, with better tracking errors, smoother control and reduced power consumption.
    To the best of our knowledge, RE+AL agents are the first RL-based controllers trained in simulation to outperform a well-tuned PID controller on a real-world controls problem that is solvable with classical control.
    }
    
\end{abstract}



\maketitle

\section{Introduction}
    \begin{figure}[t!]
        \centering
        \includegraphics[width=1\textwidth]{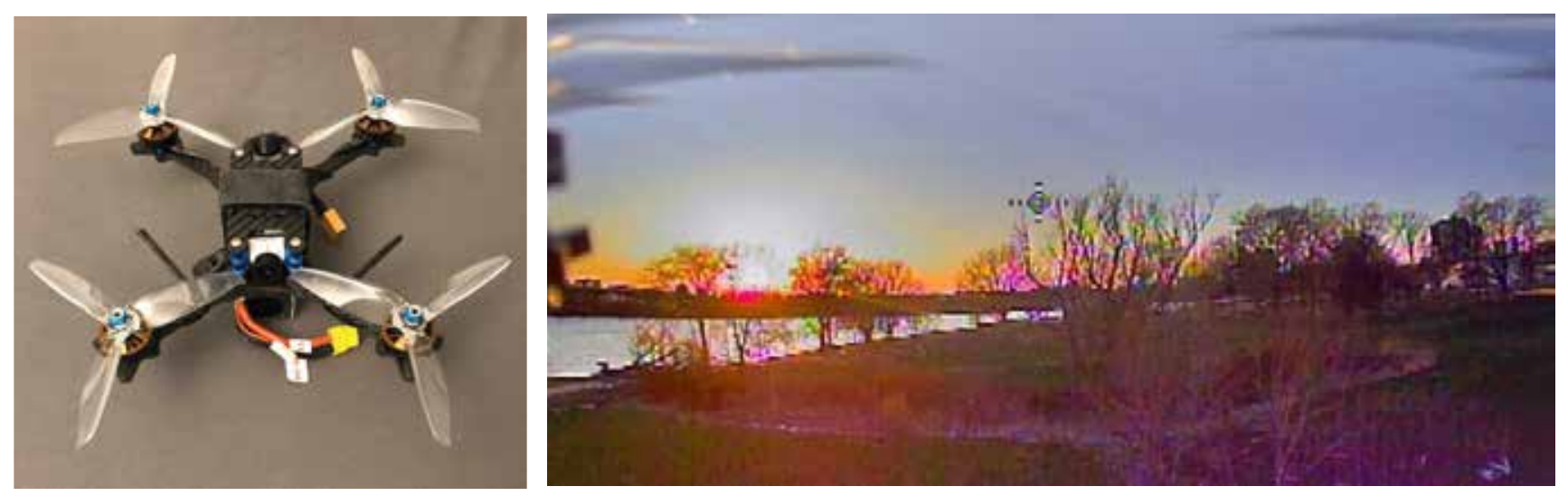}
        \caption{The NF1 platform and the pilot's view on the heads up display while flying over the treeline with with our RE+AL controller on-board.}
        \label{fig:intro_quad}
    \end{figure}

    Low-level control of physical systems such as unmanned aerial vehicles (UAVs) is usually implemented with simple PID controllers. Such controllers are straightforward to use, but cannot learn from experience or adapt to the environment. Recently, Reinforcement Learning (RL) has been applied to low-level control in embedded systems, promising a general way to train agents to map control commands to motor outputs.
    Due to safety concerns, it is not always possible or advisable to train RL agents on the platforms they are meant to be deployed on, therefore training is done in simulation.
    However, simulations are not perfect representations of reality --- this is termed as the ``reality gap''.
    The reality gap can often result in aberrant behavior when models are transferred from simulators to their real-world counterparts.
    

    A prevalent issue with learning and deploying low-level controllers trained in simulation is that the trained agents, despite strong performance in simulation, often present with a number of stability issues resulting from over-actuation (and consequent efforts to compensate)~\cite{benchmarkingRobo, Sim2multi, NFv2}.
    Despite their efforts, prior works~\cite{learning2drive, NFori, Sim2multi, Hwangbo2017ControlOA, benchmarkingRobo} found this issue difficult to mitigate reliably.
    
    {Members of our research-group previously developed Neuroflight, described in the technical report~\cite{NFv2} and described in greater detail in~\cite{NFThesis}. Neuroflight is a framework for training and deploying learned controllers on UAVs based on the GymFC~\cite{NFori} training environment.}
    While {Neuroflight} demonstrates the viability of learning low-level control for high-performance quadrotors, the authors found that the trained agents (i) did not consistently transfer with acceptable performance in real flight, despite exhibiting good performance in simulation --- with only one out of dozens of agents trained in simulation proving controllable on the real drone --- and (ii) were often unstable and non-smooth, resulting in {increased power consumption and} excessive strain and wear on the motors. 
    By analyzing their methods and results, we hypothesize that the root of the issue is the reward structure --- the basis of optimization in RL --- which is 
    highly tuned for good performance in simulation, but fails to capture the problem in a way that transfers to real-world observations and dynamics, likely due to the inherent limitations to the fidelity of simulaton.
    
    {Building on the groundwork laid by Neuroflight, o}ur work presents a systematic approach for effectively and consistently developing RL agents that can handle high-performance low-level control needs while achieving smoother outputs.
    We achieve this through a two-fold approach:
    \begin{enumerate}[(i)]
        \item We develop a new reward structure that is designed to be more intuitive and allows for faster and more repeatable training of agents that transfer reliably to a real drone.
        
        \item By carefully studying the behavior of the firmware in receiving and responding to control signals, we construct a more fitting state representation for our problem and also generate training signals that better approximate the RC-commands that might come from a pilot.
    \end{enumerate}
    We present a full-stack implementation where we adapt the simulation environment, the training, and compilation pipeline to produce flight-capable control firmware.
    All agents trained with our method offer a significant reduction in high-frequency oscillations in the motor-control signals when deployed on the real drone (when compared to the baseline method), with the peak oscillatory frequency of approximately 130Hz, down from the previous 330Hz, and at significantly lower amplitudes. Nonetheless, they maintain comparably low tracking errors, with an average error of 4.2 deg/s. 
    Our training structure also yields a 10x speedup in the time required to train transferable agents, which, on our machines, reduced training time from nearly 9 hours to under 50 minutes.
     
    \newpage 
    The remainder of the paper is organized as follows. In Section~\ref{background}, we provide a basic overview of reinforcement learning, covering the core problem structure and optimization framework. 
    Section~\ref{PriorWork} reviews related work which also tackles the problem of continuous control and details the learning pipeline for the Neuroflight framework, which we use as our primary baseline.
    Section~\ref{our-method} describes our proposed approach, RE+AL, for achieving smoother and more transferable RL agents.
    Section~\ref{EVAL} presents performance metrics for agents trained with RE+AL and compares them with the original Neuroflight method on both simulated and real-world flights.
    Finally, we conclude with Section~\ref{conclusion}.
    



 


\section{Background}\label{background}


    Reinforcement Learning represents a class of machine learning algorithms that attempt to develop an optimal states-to-actions mapping intended to maximize a numerical reward.
    The sequential decision making problem in RL is typically formulated as a Markov Decision Process (MDP), where actions influence both the immediate and eventual rewards. 
    Additionally, state transitions are assumed to obey the Markov property --- i.e. the response of the system at time~$t$ to any state and action should only be reliant on the current state~$s_t$ and action~$a_t$, which can be formalized as:
    \begin{equation*}
        P(s_{t+1}|s_t, a_t, s_{t-1}, a_{t-1},\dots,s_0,a_0) = P(s_{t+1}|s_t,a_t)
    \end{equation*}
    Extending dynamical systems theory, the problem of RL can be formalized as ``the optimal control of incompletely-known Markov decision processes"~\cite{SuttonBarto}.
    
    \begin{figure}[h!]
        \centering
        \includegraphics[width=0.63\textwidth]{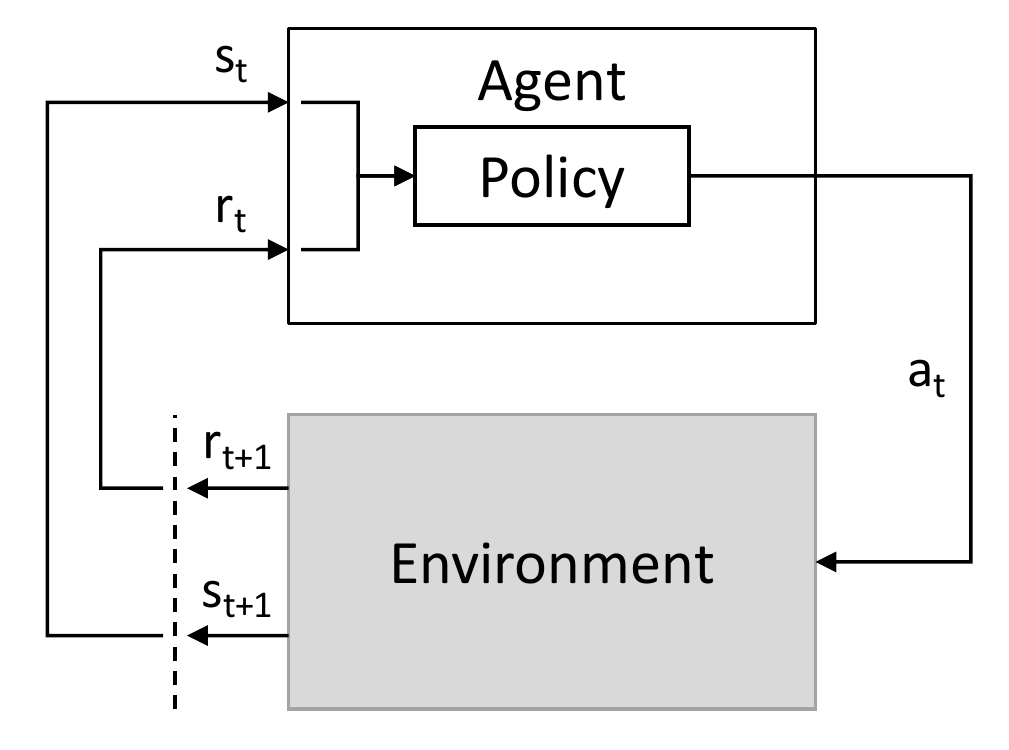}
        \caption{General structure of the reinforcement learning interaction loop.}
        \label{fig:rl_flow}
    \end{figure}
    
    There are four key elements of RL problem formulations: the (i) agent, (ii) environment, (iii) reward signal, and (iv) policy, as shown in Fig.~\ref{fig:rl_flow}.
    The learner (or agent) in RL settings are tasked with learning what to do in order to best achieve their goals through interaction with the training environment and developing an understanding of which actions to take in any given state.
    At any given time instance~$t$, the agent interacting with the environment is in state~$s_t \in S$, and can take an action~$a_t \in A(s_t)$ (an action in the valid action space $A$ of $s_t$), for which a numerical reward, $r_t$, is received.
    Typically $A(s_i) = A(s_{j\neq i})$, however, for some problems this may not always hold --- for the purposes of this paper, we assume $A(s) = A ~\forall~ s \in S$, i.e. that the action space is constant.
    The reward signal is a numerical performance measure that indicates the quality of an action in the environment, given the current state.
    The signal serves to penalize objectively bad actions while rewarding good ones to reinforce good behavior in an agent - hence `reinforcement learning'.
    The agent's goal is to maximize its expected return --- the total expected reward that it receives: 
    \begin{equation*}
        R_t = \sum_{k=0}^\infty \gamma^k r_{t+k}
    \end{equation*}
    where $\gamma \in [0,1]$ is a discount factor introduced to control the relative weight of current and future rewards in the estimation of expected total reward received.
    The policy, $\pi$, is defined as a mapping of states to probability of selecting an action given the state - i.e. $\pi(s) = \{P(a|s) ~|~ a \in A\}$.

    RL seeks to develop a policy $\pi_{*}$ that maximizes the value function of a state, $v_{\pi}(s)$ or equivalently the action-value function $q_{\pi}(s,a)$ for agents acting under a policy $\pi$:
    \begin{align*}
        \pi_{*} &= \text{argmax}_{\pi}\ \sum_{s} v_{\pi}(s) \\
                &= \text{argmax}_{\pi}\ \sum_{s} q_{\pi}(s,a^{*})\ \ \text{s.t.} \ a^{*} = \text{argmax}_a q_{\pi}(s,a)
    \end{align*}
    where
    \begin{gather*}
        v_{\pi}(s) = \mathbb{E}_{\pi}\left[ R_t | s_t=s\right] \\
        q_{\pi}(s,a) = \mathbb{E}_{\pi}[R_t|s_t = s, a_t = a]
    \end{gather*}
    It is important to note however that, while the RL agents seek to maximize the total expected reward, this is equivalently achieved (and easier optimized) when focusing on maximizing the total expected reward of any individual action, and thus the focus of RL agents is to learn to take the best action possible for any given input state.
    
    A number of tools have been developed in order to tackle the optimal policy problem.
    Policies may be optimized by on- or off-policy techniques.
    On-policy RL utilizes the same policies for exploring the state and action space of an environment as they do in optimization. 
    Conversely off-policy RL utilize a derivative of the target policy (the policy being optimized) for exploration, thus allowing for the use of distributed deployment and batch optimization techniques.
    For low-dimensional state spaces with limited actions, the optimal policy can be solved using Monte Carlo methods or approximately solved (in a less data-intensive way) using temporal difference (TD) learning techniques such as SARSA~\cite{SARSA}, Q-learning~\cite{Qlearning}, or double Q learning~\cite{DoubleQ}, with TD-learning techniques having been demonstrated as being optimal in the limit~\cite{SuttonBarto}.
    Additionally, policy gradient methods have been developed, such as REINFORCE~\cite{REINFORCE}, which represent the policy in a parametric form and utilizes gradient ascent techniques to improved the policy incrementally.
    Policy gradient techniques form the basis of many of the deep reinforcement learning techniques which have gained popularity in recent years~\cite{sutton2000policy, PPO, TRPO, dqn}.
    One of the key issues that deep RL helps addressing, which classical RL was ill-equipped for, is the application of RL techniques to continuous control.

\paragraph{\bf Deep Reinforcement Learning}\label{RL-deep}

    Applied broadly, policy gradient methods define policies~$\pi_{\theta}$ by a set of parameters,~$\theta$, and apply iterative gradient ascent to maximize policy performance.
    Given $J(\theta) \propto \tau_{\theta}$, where $J$ is a performance measure on $\theta$ --- which in RL is typically proportional of the estimated temporal-difference $\tau_{\theta}$ --- we have:
    \begin{align*}
       \tau_{\theta} &= R_t - v_{\pi_\theta}(s_t) \\
            &\approx r_t + \gamma \text{max}_{a'} q_{\pi_\theta}(s_{t+1},a') - \text{max}_{a''} q_{\pi_\theta}(s_t,a'').
    \end{align*}
    
    Parameters in policy gradient algorithms are updated proportionally to the gradient on the policy performance with respect to parameters, $\nabla J(\theta)$:
    $$ \theta_{t+1} = \theta_t + \alpha \nabla J(\theta_t) $$
    where $\alpha$ modulates the rate of change of $\theta$.
    
    Deep reinforcement learning methods extend classical policy gradient methods by representing RL policies with artificial neural networks. 
    The networks are in turn parameterized by their internal network weights and biases.
    This innovation allowed RL to capitalize upon the significant representative~capacity of neural networks for modeling arbitrary functions to tackle more challenging tasks, with works in the field demonstrating the ability to train agents to play video games, solve various controls problems and even besting humans in complex games such as chess and go~\cite{alphago,alphazero}, and even achieve `master' level play in games like Starcraft~\cite{alphastar}.
    
    While initial developments in deep RL were built to work with discrete action spaces, extensions were made to deep RL architectures to allow for the development of RL algorithms for learning action policies in continuous state and action spaces.
    This change to the policy network structure allowed for RL techniques to be developed and deployed on continuous control tasks and has been demonstrated to have a wide range of applications --- most notably a number of complex controls tasks where developing a controller following classic controls theory is highly non-trivial~\cite{DDPG,PPO,TRPO,benchmarkingRL,benchmarkingRobo}.

\section{Prior Work --- Learning Quadrotor Control}\label{PriorWork}
    
    One significant issue with deep learning techniques, or indeed machine learning techniques, is that the learned models could over-fit to the training domain.
    This has been observed to be a significant problem in deep RL, with agents often memorizing correct behavior in a given state, as opposed to learning a systematic model that would allow it to generalize to novel inputs~\cite{Sim2Real, Overfitting, Overfitting2}.
    In problems of control, this leads to noticeable aberrant behavior when agents are transferred between domains~\cite{NFori, NFv2, Sim2multi, benchmarkingRobo}, which can even cause mechanical failures~\cite{Sim2multi, benchmarkingRobo}.
    
    
    A number of recent works address the specific problem of quadrotor flight control with controllers trained with RL.
    Molchanov et al.~\cite{Sim2multi} train neural networks policies for low-level attitude control using the Proximal Policy Optimization (PPO)~\cite{PPO} algorithm.
    In their work, they define the RL cost function, i.e. their reward signal, as a weighted sum over position, velocity, and angular velocity errors, with additional penalties on the drone's acceleration and orientation.
    They investigate the sensitivity of their RL agents as a function of the weight parameters and provide a performance metric over tracking errors and oscillations in the quadrotor's frame. They note that, while some agents allowed for stable flight, others might result in issues ranging from visible oscillations to an inability to take off.
    They do not however investigate the underlying causes of the oscillations, nor do they report on control signal behavior during tests, and their effect on the motors.
    Related work by Hwangbo et al.~\cite{Hwangbo2017ControlOA}, also presents a full pipeline for training RL agents for quadrotor flight control.
    Similar to Molchanov et al., they construct their reward signal additively over state tracking errors.
    Unlike Molchanov et al., to achieve stable flight, they employ a varied exploration strategy for exploring the state-space.
    They also make stronger assumptions on prior knowledge of the quadrotor dynamics and augment training with a re-tuned Proportional Differential (PD) controller.
    While the authors comment on the stability or lack thereof of their quadrotor in various flight conditions, they do not offer a quantitative analysis of the observed phenomena.
    {Our work builds upon the preliminary results we obtained in the Neuroflight framework, for which a technical report is available in~\citet{NFv2}. Neuroflight is also described in greater detail in~\cite{NFThesis}. The tools developed through the Neuroflight work are summarized in Sections~\ref{NFbasics}~and~\ref{WilThesis}.}

\subsection{Neuroflight}\label{NFbasics}
\begin{figure}[h]
    \centering
    \includegraphics[width=0.7\textwidth]{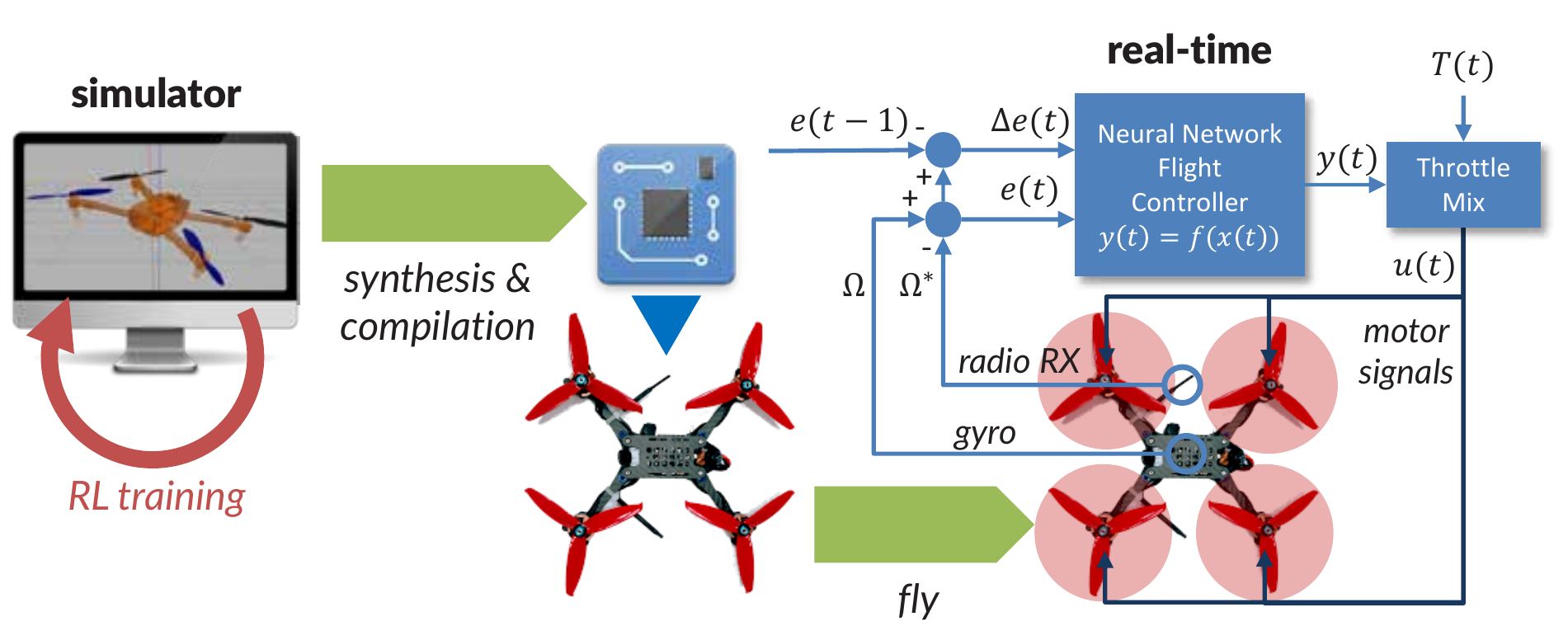}
    \caption{An overview of the Neuroflight framework~\cite{NFv2}}
    \label{fig:nf_framework}
\end{figure}

    The Neuroflight framework introduced by Koch et al.~\cite{NFv2} was designed to be a full-stack solution for training, testing and compiling neural-network-based flight controllers into firmware that can be flashed on off-the-shelf micro-controllers (see Fig.~\ref{fig:nf_framework} for an overview).
    The training environment, called GymFC and introduced in~\cite{NFori}, is a simulator built using Gazebo~\cite{gazebo}, to provide an OpenAI Gym-compatible~\cite{GYM} interface for training RL agents.
    GymFC allows simulated controllers to interact with simulated quadrotors safely in a virtual environment, making it a good training and testbed for our agents.
    With their original work, Koch et al. provide a feasibility study on the use of controllers trained using deep RL techniques for deployment on real drones.
    They benchmarked a number of the then state-of-the-art deep RL algorithms on the task of low-level drone attitude control and, like Molchanov et al.~\cite{Sim2multi}, were able to successfully demonstrate the viability of the PPO algorithm on this task.
    They further establish that agents trained with PPO were able to attain faster convergence to rotational angular rates taken from real flight tests than a tuned PID controller both in the real world and in simulation.
    
    

\subsection{Neuroflight tuning and improvements}\label{WilThesis}

    As identified by Koch et al.~\cite{NFori,NFv2}, one of the key issues with the control policies learned by the RL agents with the Neuroflight framework is the lack of smoothness and stability in the control output.
    These instabilities were previously observed both in simulated tests as well as when transferred to the quadrotors for real-world flight.
    In \revision{a technical document extending} their prior work, Koch et al.~\cite{NFThesis} further developed the simulation and RL training sub-systems in an effort to improve set-point tracking and flight stability.
    They focused primarily on improving model fidelity for the quadrotor within the simulator, and tuning the neural network construction and reward-signal for the deep RL algorithm employed.
    We briefly discuss these improvements to help contextualize our work, which aims to address some of the \revision{key remaining} issues in the Neuroflight policy training framework.
    \revision{We also provide an analysis on how remaining model limitations, even in the higher fidelity model, contribute to a still significant reality gap in Section~\ref{realityGap}.}
    
    While initial versions of the simulated quadrotor training environment were built with a generic off-the-shelf quadrotor model, Koch et al. developed a higher-fidelity model of the target quadrotor platform.
    By building a custom Gazebo model for the simulator which better represented the shape, weight, dimensions and thrust forces of the NF1 quadrotor, Koch et al. are able to more accurately model both the quadrotor construction and the motors' properties.
    To better simulate the physical forces acting on the quadrotor, Koch et al. also changed the internal physics engine in Gazebo to work with DART~\cite{DART} instead of the standard ODE~\cite{ODE} engine typically employed by Gazebo.
    In our work, described in Section~\ref{our-method}, we employ these same improvements to the simulator.
    
    The problem of control stability also often arises because stability is not a criterion typically optimized for in RL.
    To address these issues, Koch et al. engineer their reward signals to encourage a smooth control response.
    Their new reward signal is comprised of five components: 
    (i) a penalty for high motor acceleration, $r_a$; 
    (ii) an in-band-reward for maintaining low motor activation while staying within an error threshold, $r_b$; 
    (iii) a penalty proportional to the state error of the quadrotor, $r_e$;
    (iv) a penalty for attempting to over-actuate the motors, $r_o$;
    (v) a penalty for not actuating the motors, $r_n$.
    \newpage
    \noindent 
    These five reward components are composed additively to give the final state reward at time $t$, $r_t$:
    
    \begin{equation}
        r_t = r_a + r_b + r_e + r_o + r_n \label{WillAdd}
    \end{equation}
    with the five individual reward components computed as:
    \begin{align}
        r_a &= -100 \cdot \max_i\left(|y_{t}^{(i)} - y_{t-1}^{(i)}|\right) \label{WilAcc}  \\
        r_b &= 1000 \cdot \left(1 - \frac{1}{4}\sum_i y_t^{(i)}\right)\mathds{1}_{{\bf e}_t < \epsilon} \label{WilInBand}\\
        r_e &= ||{\bf e}_{t-1}|| - ||{\bf e}_{t}|| \label{WilError}  \\
        r_o &= -1\times10^9 \cdot \sum_i \max(y_{t}^{(i)} - 1, 0) \label{WilOversat}  \\
        r_n &= -1\times10^9 \cdot \mathds{1}_{||{\bf \bar{\phi}}_t|| > 0} \cdot \sum_i \mathds{1}_{y_t^{(i)} = 0} \label{WilZeroAct}
    \end{align}
    where $\mathds{1}_z$ is an indicator function on criterion $z$, $y^{(i)}_t \in [0,1]$ is the current motor-control signal output by the RL agent for each motor $i \in \{1,2,3,4\}$, ${\bf \bar{\phi}}_t$ is the desired angular roll, pitch and yaw velocity set-point, ${\bf \phi}_t$ is the current angular roll, pitch and yaw velocity, and ${\bf e}_t = {\bf \bar{\phi}}_t - {\bf \phi}_t$.
    While this reward signal structure was shown to consistently achieve smooth control in simulation, the performance of the agents in real world flight was observed to still present with high-frequency oscillations in the motor control signals, contributing to significant motor heating and power drain.

\section{RE+AL: REinforcement-based transferable Agents through Learning}\label{our-method}
    
    Prior work in the field has established the viability of deep RL in continuous controls task and the works of Koch et al.~\cite{NFThesis} provided us with a baseline upon which to build.
    There are a number of crucial pending problems with controllers trained on the Neuroflight pipeline, namely the lack of reproducibility and failure to preserve smooth control in transfer.
    We posit that the main issue is the over-engineering of the reward signal \revision{which results in less stable optima, contributing to agents being highly sensitive to network dynamics and brittle on the domain shift from simulation to reality}.
    {To combat this issue, we developed a new reward scheme to better reflect desired network behavior, without being as closely tied to the environment dynamics, while also modifying the training environment to encourage a better exploration of the state-space.}
    
    The reward structure introduced by Koch et al., which was summarized in Section~\ref{WilThesis}, allowed for the successful training of PPO agents. However we found that the reward structure was far too fine-tuned for the very specific network architecture and simulator design, with minor changes in either causing the agents' training to break down. 
    Despite the highly engineered rewards, the final agent(s) did not reliably transfer to the real drone.
    To mitigate this, the authors had to select potentially flight-capable agents manually through trial-and-error with basic flight tests.
    Furthermore, even the best agent presented by Koch et al.~\cite{NFv2, NFThesis} was noisy in its control actuation, with high-frequency oscillations, resulting in significant motor heating and power consumption.
    We believe this reflects a reward structure fundamentally ill-suited to represent the control problem at hand.
    \revision{As our results show in Section~\ref{EVAL}, there exists a reality gap between simulated and real dynamics but is not so large, on the time scales that the controller operates at, that it should result in a catastrophic breakdown of learned behavior, as we observe in the instability of policy-transfer of Neuroflight agents.
    Also evident from the training data presented by Koch et al. is the high variance in training performance (Figs. 5.16 and 5.22 in~\cite{NFThesis}).
    This strongly suggests that the rewards used in the existing Neuroflight pipeline do not appropriately reflect the quality of an agent's learned behavior in a transferable way.
    }
    
    In the remainder of this section, we will discuss the systematic process by which we developed `REinforcement-based transferable Agents through Learning', or RE+AL, a new regimen for training control policies.
     While we demonstrate the viability of our methods \revision{primarily} on a quadrotor drone, the underlying principles of our design, \revision{the core of which is the newly proposed reward structure}, should apply to a broader spectrum of control problems, \revision{as discussed in Section~\ref{openAIGymEvals}}.

\subsection{Constructing and Composing Multi-Objective Rewards}\label{ourRewards}
    \revision{
    The standard RL formulation for training a policy generally requires a scalar reward signal to indicate the quality of policy's performance on a task.
    In cases where rewards may be broken down into multiple components, the reward signal provided to the RL agent during training needs to be scalarized.
    An interesting note about multi-objective RL however is that most non-trivial Reinforcement Learning problems are often multi-objective in nature.  
    Consider, for example, the Pendulum-v0 environment from OpenAI's Gym~\cite{GYM} benchmarks, one of the `simplest' problems in the set of benchmarks --- it too balances costs on joint angles, velocity and acceleration.
    There is however no clear rule on how scalaraziation ought to be handled when considering multiple reward components.
    The most common composition operator we found is additive, as used by Koch et al. (see Equation~\ref{WillAdd}) and often in the literature \cite{Sim2multi, Hwangbo2017ControlOA, learning2drive, dynamicweights, shelton2001balancing}, and can be generalized for a K-dimensional reward vector, $r \in \mathbb{R}^K$, as follows:
    \begin{equation}
        r_t = \sum_k^K w_k r_k \label{additive}
    \end{equation}
    }
    
    \vspace{-\baselineskip}
    \revision{
    Composing rewards additively in this manner can result in a number of complications. 
    Specific weight assignments may lead to very different states and policies being explored, drastically affecting the final learned policy (discussed further in Section~\ref{openAIGymEvals}). 
    Rewards need to be carefully tuned to avoid introducing undesirable local minima in the optimization space where agents exploit one component of the reward composition at the cost of others.
    This can be especially problematic when different reward components compete and interfere destructively with each other.
    The fundamental problem with additive composition is that it does not impose a strong requirement on all objectives being satisfied and allows for some objectives to go unsatisfied if other objectives can compensate for them.
    This means that policy optimization may learn to ignore some objectives.
    We desire instead an operator encouraging the satisfaction of \emph{all} reward objectives, i.e. a logical AND operator.
    }
    
    \revision{
    We propose multiplicative composition as an alternative to additive composition.  
    For a vector of normalized reward components, we propose taking the product of all the individual reward components i.e. using a product t-norm operator, which acts as a smooth generalization of the AND operator and is often used in Fuzzy logic~\cite{Hajek1998}.
    With multiplicative composition and normalized reward components, the importance of each component reward signal is preserved.
    If any one reward component falls too low, the overall reward signal is lowered regardless of the quality of the other components. 
    In other words, the agent is not allowed to offset its poor performance on one objective by being better on another.
    This property is essential if the objectives defined are competitive. 
    This type of composition encourages agents to learn to jointly optimize across all the reward components.
    It also has the advantage of being scale-invariant as the relative scales of the different reward components do not cause any of them to overshadow the others.
    It is however sensitive to translation, where clipping a reward can change its ``importance".
    We can use this property to our advantage to specify which rewards we value more than others.
    Instead of simply taking the product across the rewards however, we take their geometric mean \cite{Fleming1986HowNT} $g(r)$ defined as:
    \begin{equation}
        r_t = g(r) = \left(\prod^K_k \min(1,r_k+\epsilon)\right)^{K^{-1}}
    \end{equation}
    For normalized rewards in the range of $[0,1]$, taking the $K$-th root preserves the scale of the rewards and prevents the final scalar value from vanishing the more rewards are added.
    We also introduce a small $\epsilon$ to ensure that all components are non-zero to mitigate loss of information from non-zero rewards in the case of a zero-reward-component.
    }

    \revision{
    To the best of our knowledge, we are the first to propose multiplicative reward composition as a deliberate design choice in representing reward signals for better learning in RL.
    While this paper primarily addresses the utility of this multiplicative rewards operator in the context of training RL agents for aerobatic quadrotor flight control, we believe that this form of rewards composition can have broader utility in improving the consistency of trained policies.
    We demonstrate this in Section~\ref{openAIGymEvals}, which analyzes the behavior of multiple OpenAI Gym benchmark environments when modified to use multiplicative reward scalarization instead of additive.
    Our results show that multiplicative composition allows more simplified reward composition with little to no tuning to inform successful training of RL agents.
    The performance of agents trained on multiplicative composition presents with significantly reduced variance while achieving comparable or better rewards when compared to their counterparts trained under additive composition.
    }

    
   

\subsection{Reward Composition for Quadrotor Attitude Control}\label{rewards}

    
    
    When considering the problem of training a low-level controller for drones, we require three key performance traits: (i) a low tracking error --- the agents need to respond accurately to input control signals; (ii) smoothness --- the control signal should have minimal unnecessary actuation to preserve the life of the hardware components; and (iii) maintaining a minimum output --- all motors need to output at least enough power to keep the drone airborne.
    
    Koch et al. identify and attempt to account for these same performance metrics.
    Equation~\ref{WilError} attempts to encourage low tracking error.
    Equation~\ref{WilAcc} penalizes acceleration in the motor, thus promoting smooth control signals.
    Equations~\ref{WilInBand},~\ref{WilOversat}, and~\ref{WilZeroAct} provide feedback on the quality of the motor signal and drive the policy to maintain a low, non-zero level of actuation.
    The main problem with these reward signals however is that they are both adversarial and on vastly different scales --- with some reward signals being several orders of magnitude larger than others --- and this often results in destructive interference in behavior optimization~\cite{amodei2016concrete}.

    
    
    To work with our multiplicative composition, we define a new set of reward signals to reflect \revision{important} controller performance traits. 
    We begin by reflecting each performance trait as a penalty on each of the performance criteria:
    \begin{enumerate}[(i)]
        \item Error penalty defined by the difference between the set point and the actual angular rate of the drone
          \begin{equation}
              p_s = || \phi_t - \bar\phi_t ||_4;
          \end{equation}
        \item Smoothness penalty on each motor, $i$, which penalizes change in motor usage
          \begin{equation}
              p_c^{(i)} = |y^{(i)}_t - y^{(i)}_{t-1}|;
            \label{eq:r_smooth}
          \end{equation}
        \item A thrust penalty on each motor, $i$, which is intended to keep motor usage at~$\mu$, the average power used for maintaining stable thrust for maintaining altitude, while penalizing it for over actuating
          \begin{equation}
              p_u^{(i)} = |y^{(i)}_t - \mu|.
          \end{equation}
        {The networks we train are designed specifically for attitude control.
        Thrust control is handled separately by a thrust mixer which mixes a pilot's thrust input with the motor actuation required for the desired rotational velocities.
        We observed that an average 34\% thrust was required for stable hovering when flying with a tuned PID controller.
        Treating this as a baseline for the minimum thrust required during flight, we utilized a value $\mu = 0.34$ in training to better match the simulated dynamics with the expected minimum thrust during flight.}
    \end{enumerate}
    The penalties are cast as positive rewards,~$p^+$, by reflecting it around 0 and clipping it to $[0,1]$, i.e.:
    \begin{equation}
        p^+(p) = \min(1,\max(0,1-p)).
    \end{equation}
    This results in the reward components being defined as:
    \begin{align}
        r_s &= p^+(p_s/\beta) \\
        r_u &= g(p^+(p_u)) \\ 
        r_c &= g(p^+(p_c))
    \end{align}
    where $p_s$ is scaled and bounded by $\beta$, such that everything above $\beta$ deg/s would be considered a 0.
    In practice, we observed that $\beta=300$ served as a good upper bound for our drone in training.
    Finally, the composite reward signal, $r_t$, used in training is composed as:
    \begin{equation}
        r_t = g([r_s, r_u, r_c]).
    \end{equation}
    
        
    
\subsection{State space}~\label{statespace}
    
    Given knowledge of the reward signals, it is important to define the state space under which the agent operates in a way that contains all the relevant information that contributes to the reward signals.\\\\
    Koch et al.~\cite{NFv2,NFThesis} originally composed their state vector, $s_t$ as:
    \begin{equation*}
        s_t = [{\bf e}_t, {\bf e}_t - {\bf e}_{t-1}]
    \end{equation*}
    to capture the current tracking error and change in angular velocity, i.e. the angular acceleration, of the drone.\\\\
    We instead define our state vector, $s_t$ as:
    \revision{
    \begin{equation}
        s_t = [{\bf e}_t, \phi_t, \phi_t - \phi_{t-1}, y_{t-1}].
    \end{equation}
    }
    While compressing information about the current and desired angular velocities into a single error quantity does allow for a reduction in state complexity, it is predominantly only useful when one can assume that control responses are always the same given any error, regardless of the current underlying state of the system.
    We found that this is not the case in practice, with the required control response varying depending on the angular velocity of the drone.
    In light of this, we chose to additionally include current angular velocity and acceleration in our state space, instead of leaving the networks to learn the disambiguation on their own.
    We also chose to include the last action taken by the agent as a part of the state space.
    When penalizing an agent for having significant changes in actions taken between time steps (see Equation~\ref{eq:r_smooth}), it is important for the agent to be aware of what action was last taken in order to effectively learn to remain smooth.
    
\subsection{Goal generation}~\label{goalgen}
    
    The original Neuroflight training framework provided training goal signals, i.e. the desired set-point angular velocities for agents, as a sequence of step inputs.
    Over a single training episode\footnote{An episode is defined as a contiguous sequence of interactions between the agent and the environment across several simulated seconds.}, the goal was changed once from 0 deg/s to a randomly selected value within the quadrotor's flight envelope, held constant for a few seconds and returned to zero.
    This was not a good reflection of the types of signals received during actual flight, as human pilots would typically perform continuous maneuvers, rarely holding at a specific target velocity. 
    We further hypothesized that the long periods of consistent signal levels might be conditioning the policy's estimations of state transition probabilities to be biased towards minimal change.
    
    In order to get more realistic set-point changes in target angular velocities during training, we opted to procedurally generate set-points according to a Perlin noise\cite{10.1145/325165.325247} function.
    Goal generation built on Perlin noise allows us to have constantly changing random set points but preserves smoothness between subsequent timesteps.
    Furthermore, agents are able to explore more of the state space in individual episodes, instead of being conditioned for stable goal inputs.
    In our experience, this contributed to agents learning faster and seemingly to generalize better, likely due to not over-estimating the transition probabilities between self-same states.
    
    We further augment the base Perlin noise function by multiplying it with a slower Perlin noise function so that the environment can be conditioned to expose the drone to a balance of steady as well as more aggressive maneuvers.
    To aid in the agents' ability to generalize over non-continuous inputs. 
    
    For a Perlin noise function, $\mathcal{P}(t, o)$ defined on time, $t$ and the number of desired octaves, $o$, we define our set-point generation function, $\bar\phi(t)$, as:
    \begin{equation}
        \bar\phi(t) = \mathcal{P}(t, 4)\times \mathcal{P}(t, 1)^2  
    \end{equation}
    The first Perlin component, with an octave count of $4$, generates a sequence of values whose rate of change is neither too smooth nor noisy. 
    The second component biases the set-point to be closer to zero, which is intended to help the function to more closely resemble real-life controls. Indeed, pilots do not constantly change their command inputs, nor do they do it at a constant rate. 
    
    During training episodes, we monitor the agent's tracking error and terminate episodes early if errors grow too large.
    This is done both to indicate to agents when their behavior is unacceptably bad, and also to avoid wasting training time on unrecoverable states.
    
    \begin{figure}[h!]
        \centering
        \includegraphics[width=.8\textwidth]{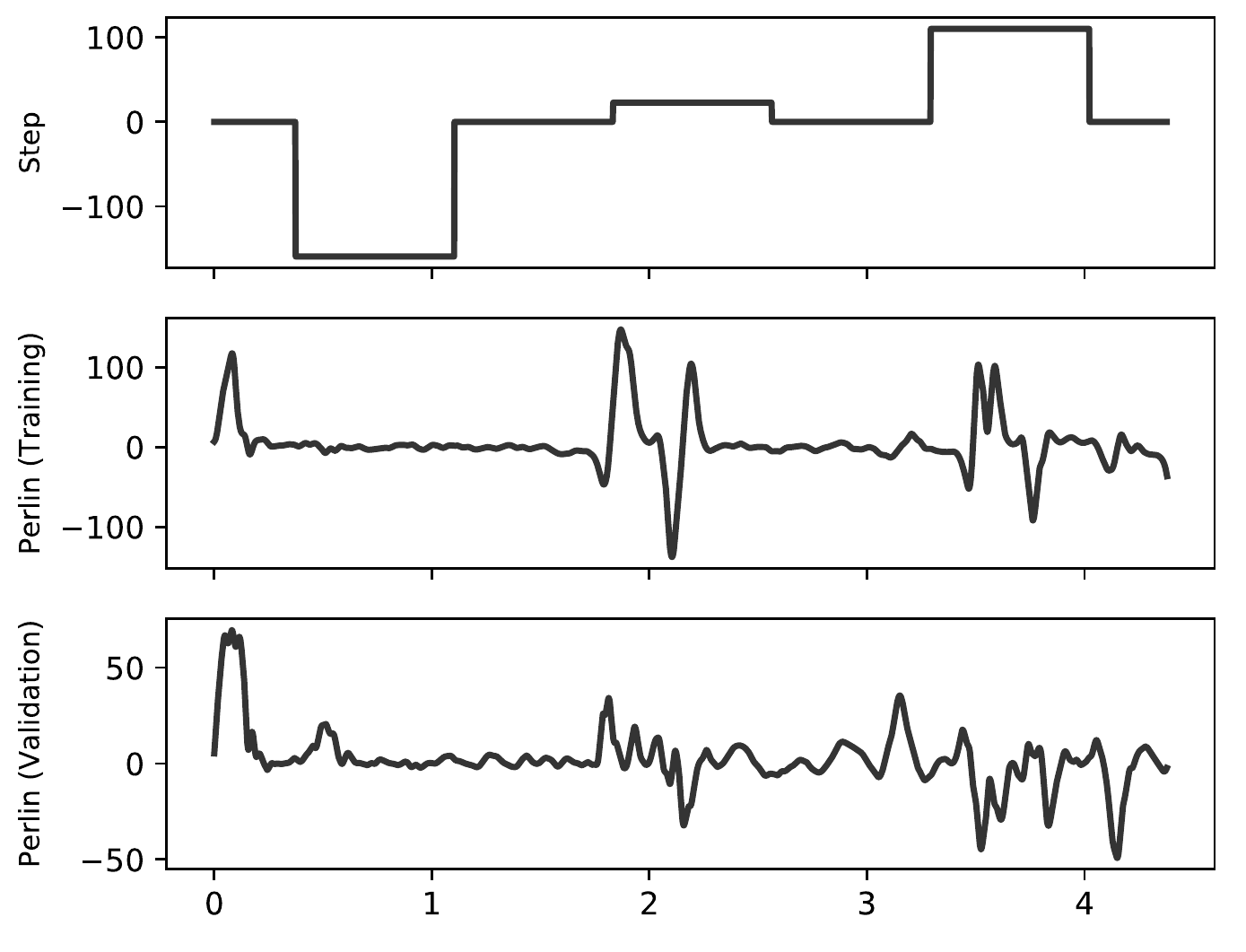}
        \caption{Neuroflight's step set-point generator compared to our Perlin based one -- We make use of a more aggressive procedural generator during training to encourage agents to explore more of the state space, and validate on a gentler signal, one that is meant to reflect more typical inputs from a pilot.}
        \label{fig:perlin_vs_step}
    \end{figure}

\subsection{Specific Network Architecture}\label{netarch}
    In this work, we use a standard PPO agent, defined by and trained with open-source code provided by OpenAI baselines~\cite{baselines}.
    RL algorithms are often sensitive to the specific dynamics of the environments they are trained in~\cite{benchmarkingRobo,benchmarkingRL}.
    We specifically use PPO as it has already been proven to be a viable training algorithm for this specific problem and drone.
    The policy network we train is a fully-connected network with 2 hidden layers, and 64 artificial neurons per hidden layer.
    The output layer of the policy network has 4 neurons --- one for each of the motor command signals.

\subsection{Compilation Toolchain}\label{compilation}
    
    At the end of training, the TensorFlow~\cite{abadi2016tensorflow} graph representing the trained agent is frozen and saved as a TensorFlow checkpoint.
    The checkpoint is then optimized and compiled using TFcompile, a compilation tool offered by TensorFlow, into a self-contained executable, which is then linked with a custom version of the Betaflight firmware~\cite{betaflight-homepage}.
    Finally, the firmware is compiled to generate a hex file ready to be flashed on the quadrotor.
    Typically, Betaflight is configured to work with a Proportional Integral Differential (PID) controller but we replace this control block with a call to our trained neural network agents. Aside from the replaced controller, the Betaflight firmware offers routines for (i)~reading the remote control commands for the desired angular velocity, (ii)~reading and filtering the angular velocity from the on-board gyroscope sensor, and (iii)~writing the motor control signals to the electronic speed controller (ESC).
    
    Due to the relatively small size of the neural networks used, the final optimized neural network is 12KB large and operates at approximately 730Hz on an ARM Cortex-M4 processor clocked at 216MHz.


\section{Evaluation}\label{EVAL}

    Our primary goal is to be able to reliably and consistently train agents for low-level attitude control. We therefore define our evaluation criteria predominantly on the quality of controller transfer to the real drone as observed in real-world flight tests.
    As in Section~\ref{ourRewards}, performance metrics are defined on control input tracking error, ${\bf e}_t = \phi_t - \bar\phi_t$ and smoothness, $\Delta y_t = y_t - y_{t-1}$.
    Before deeming any agent to be potentially flight-worthy, however, it is important to verify their performance in training and simulation.
    
    In this section, we discuss the performance trends achieved in simulation during training and the performance of the controllers transferred to the physical drone.
    Our experiments show that agents trained using our pipeline consistently yield good tracking performance both in simulation and in the actual test flights. While tracking performance is comparable to the baseline established by \citet{NFv2}, our controllers present with significantly less strain on the motors thanks to the absence of high-frequency control oscillations.
    
    We use the techniques proposed by Koch et al.~\cite{NFThesis} as the baseline for our evaluation by comparing our agents against the best agents produced by their methods.
    {We also compare our agents against PID controllers tuned by the Ziegler-Nichols method~\cite{ZieglerNichols} to further demonstrate the relative efficacy of RE+AL in developing good controllers.
    The PID tuning was performed independently for simulated and real control in recognition of the differences in simulated dynamics from real-world physics. 
    This also provides the most fair comparison of agents trained by our method against controllers specifically tuned for optimality in their respective domains.}
    
\subsection{Training Performance}
    \begin{figure}[t]
        \centering
        \includegraphics[width=0.95\textwidth]{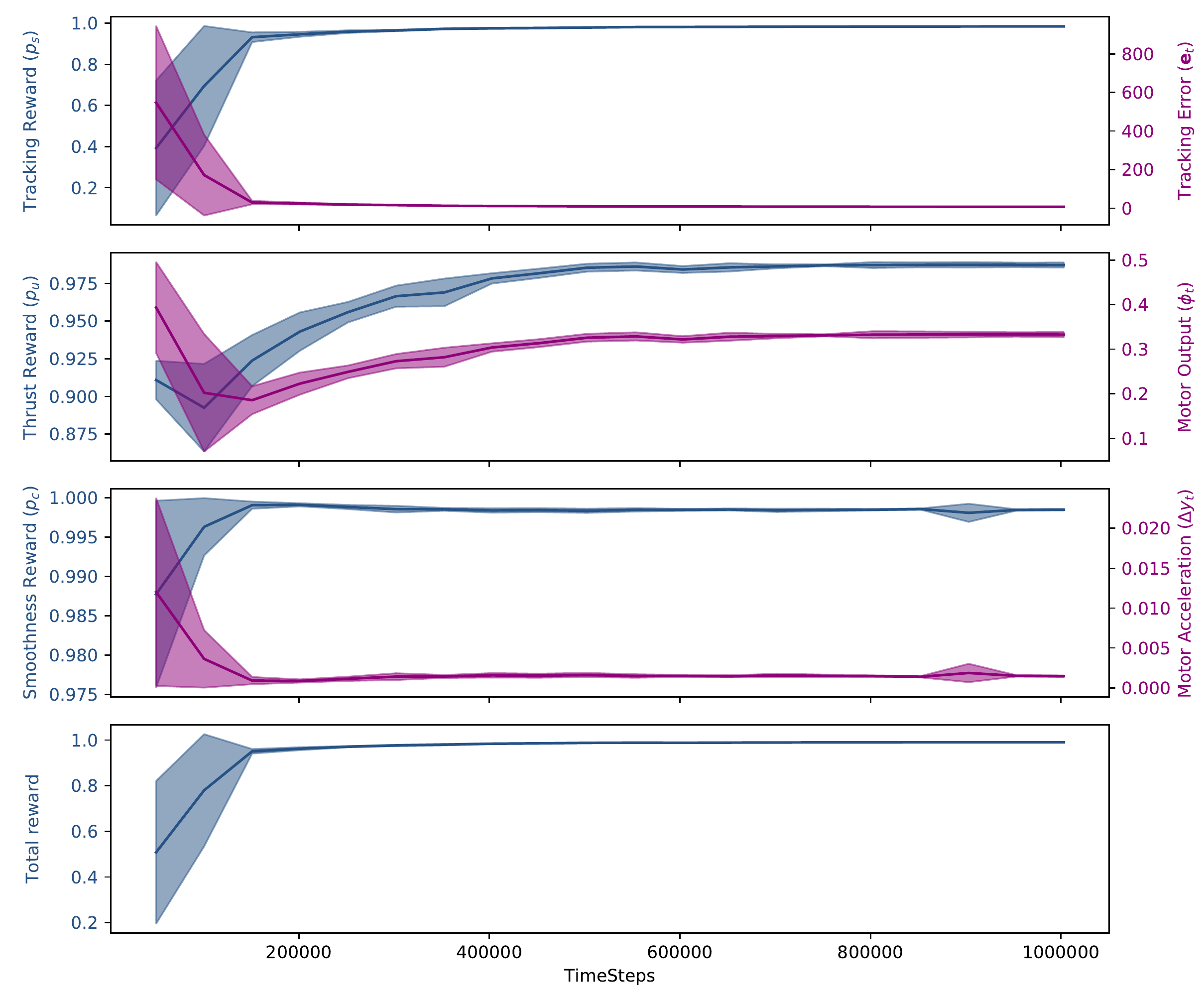}
        \caption{{Tracking RE+AL training progress - The performance of agents in simulation begins to plateau after approximately 200,000 timesteps. Despite this however, there are still minor improvements to the agents' behaviors that are learned. We practice a form of early stopping to terminate training after the average reward in simulation falls below a threshold in the interest of decreasing the number of samples required before reaching what we consider a high performance agent. For our drone, this seemed to happen by approximately 1 million timesteps --- testing agents trained for longer periods of time revealed that, while they achieved better performance in simulation, their performance in transfer was often worse, we suspect this is because the agents overfit to the specific dynamics of the simulation. {We also show how the behavior of our trained agents develops to satisfy the three performance criteria outlined in Section~\ref{ourRewards} and their respective rewards.}}}
        \label{fig:training_progress}
    \end{figure}
    
    We begin by analyzing the trends observed in training in order to verify the performance and reproduciblity of agents in simulated flights.
    Throughout the course of training, checkpoints are logged at regular intervals of 50,000 training steps (corresponding to 50,000 interactions between the agent and the simulated environment).
    The performance of the agents at each checkpoint is then validated in simulated flight sequences while the errors, rewards and simulated motor behavior are logged.
    Each training cycle begins with a randomly initialized training policy and a different random seed.
    
    Fig.~\ref{fig:training_progress} shows the typical trends in training for our RL agents.
    Note that tracking error and motor acceleration are minimized, as expected, while average motor actuation is driven to 34\%, exactly as intended by our reward signal design presented in Section~\ref{ourRewards}. We observe that the performance of the agents in simulation tends to plateau after approximately 500,000 timesteps. We allow the agent to train slightly longer so as to further reinforce good behavior.
    We also observed that allowing agents to continue training until they reach 1,000,000 timesteps typically resulted in a late-stage improvement in tracking errors, bringing the simulated-environment tracking error below an average of 10 deg/s.
    We practice early-stopping, a process by which training of ML models is stopped even if performance appears to be improving in training, to prevent over-fitting to the training domain and improve the chances for better transfer to the test domain.
    We found that stopping the training at approximately 1 million timesteps --- i.e., just after agents typically learn to reduce tracking error in training below 10 deg/s --- offered a good balance of tracking, stability and policy transfer.
    
    The training trends of our agents contrast starkly with agents trained by Koch et al.~\revision{\cite{NFThesis} in three significant ways (compare against Figs. 5.16 and 5.22 in~\cite{NFThesis})}: (i) we achieve a 10x reduction in required training time, which in practice corresponds to training for 50 minutes on our machines as opposed to 9 hours; (ii) the performance of our agents remains relatively stable even after plateauing instead of experiencing a sudden, inexplicable drop in performance half-way through training, and (iii) as we will demonstrate in Section~\ref{EvalTest}, agents trained by our method can simply be transferred to the drone at the end of training without requiring an additional search over the checkpoints to identify `good' agents. Hence our training exhibits good repeatability.  
    
    \begin{figure}[h]
        \centering
        \includegraphics[width=0.8\textwidth]{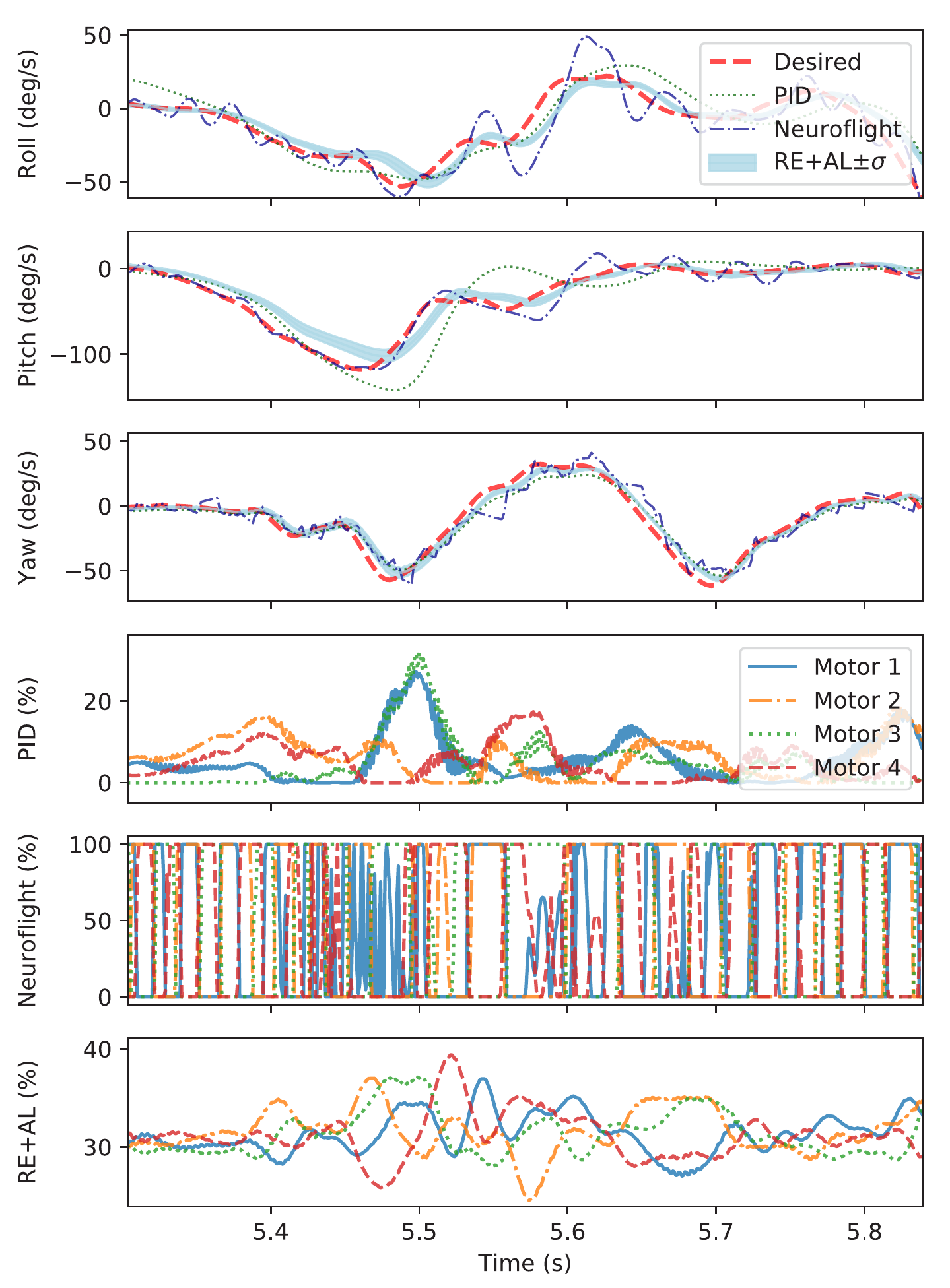}
        \vskip -0.1in
        \caption{Zooming in to analyze stability - By zooming in on the agents' behaviors over 0.5s of simulated time, we show that RE+AL control signals are very smooth and stable in simulation and our agents present with practically no oscillatory behavior.}
        \label{fig:Zoom}
    \end{figure}
    
    \begin{table*}[h]
        \caption{Comparing Tracking Errors
        \revision{RE+AL performs consistently better and with low variance on this control task than the existing baseline, Neuroflight, and even compared to a tuned PID controller.}
        }
        \label{MAE table}
        \begin{center}
        \begin{tabular}{c||c|c|c|c|c}
        \hline
        {Agent} & \multicolumn{4}{c|}{Mean Absolute Error (deg/s)}  & Current            \\ \cline{2-5} 
           &       Roll     &      Pitch   &       Yaw     &       Overall   & (Amps)   \\ \hline \hline
        \multicolumn{6}{c}{Simulated Validation} \\ \hline
        {PID}          & $10.31$      &      $17.21$      &       $6.75$      &       $11.41$   & -NA-       \\ 
        {Neuroflight}  & $8.85$      &      $\bf 6.65$      &       $6.48$      &  $7.30$   & -NA-         \\ 
        \emph{RE+AL} (ours)        & $\bf 7.00 \pm 2.37$  & $ 7.80 \pm 3.50$  & $\bf 5.76 \pm 1.51$  & $\bf 6.75 \pm 2.13$   & -NA-         \\ \hline
        \multicolumn{6}{c}{Test-Flight} \\\hline
        {PID}          & $6.68$      &      $4.54$      &       $3.80$      &       $5.01$    &  $8.07$     \\
        {Neuroflight}  & $6.26$      &      $5.42$      &       $3.87$      &       $5.19$    &  $22.87$    \\ 
        \emph{RE+AL} (ours)       & $\bf 6.01 \pm 1.80$  & $\bf 4.52 \pm 1.18$  & $\bf 2.64 \pm 0.18$  &  $\bf 4.20 \pm 1.14$ & $\bf 5.86 \pm 3.10$ \\ \hline
        \end{tabular}
        \end{center}
        
    \end{table*}

    Table~\ref{MAE table} demonstrates the high accuracy of our agents in simulation, with mean average errors of $6.75$deg/s and a relatively low, and almost imperceptible variance of $2.13$deg/s.
    These metrics further speak to the repeatibility of agents trained by our method and their apparent insensitivity to random seeds.
    
    Fig.~\ref{fig:Zoom} provides a temporal close-up on the tracking behavior of 3 independently trained agents in simulation to allow readers a better sense of how individual agents perform on simulated control tasks.
    RE+AL agents also generally exceed the set-point tracking performances of both a tuned PID and the best Neuroflight agent (provided by Koch~et~al.).

    \begin{figure}[t]
        \centering
        \vskip -0.1in
        \includegraphics[width=\textwidth]{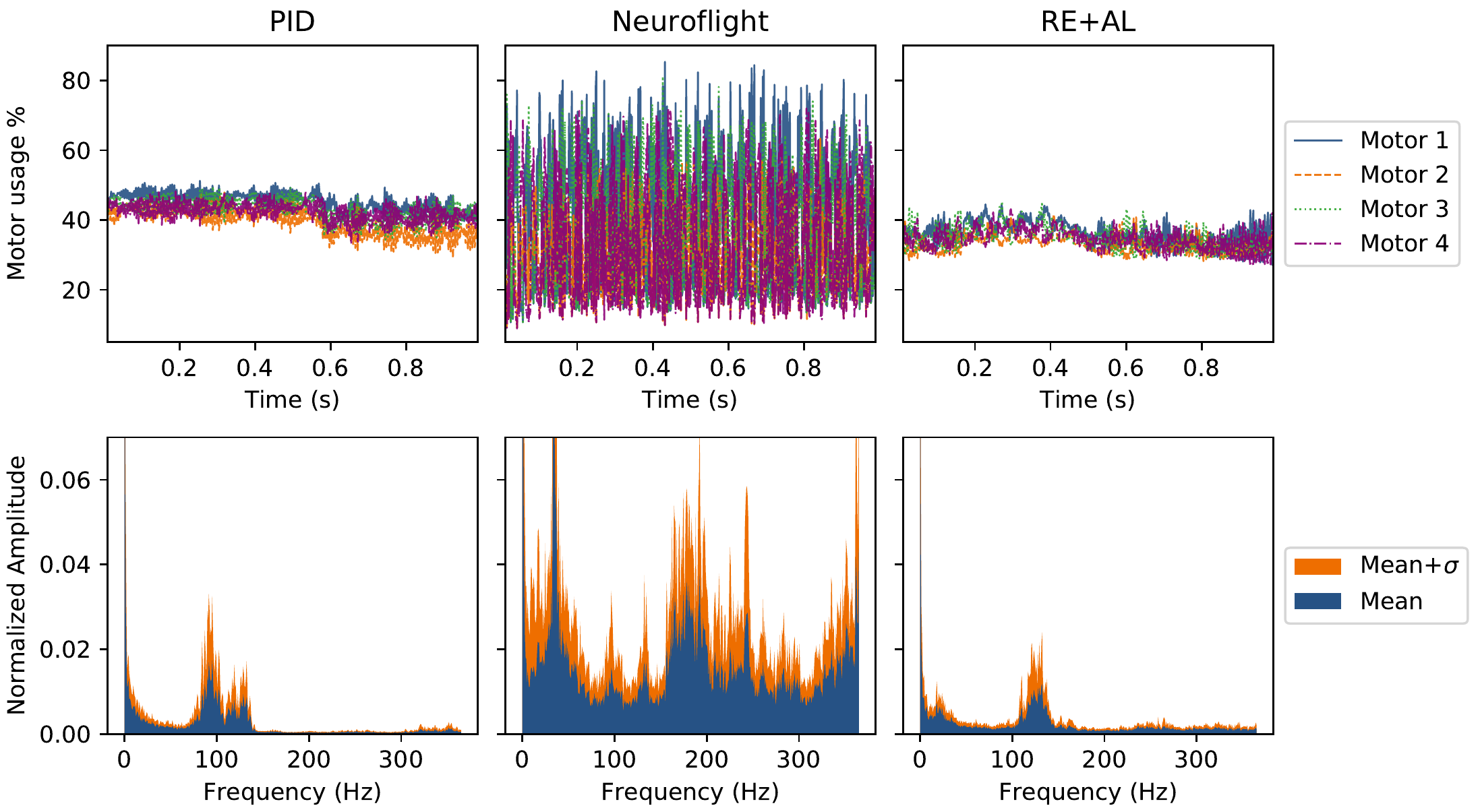}
        \vspace{-0.3cm}
        \caption{{Fourier transform of control signals - Each column shows a slice of the motor usage during similar real flight maneuvers (top), in comparison to the mean frequency spectrum resulting from a Fourier transform on the control signals of the quad-rotor's 4 motors (bottom). This visualization is presented on the same scale to highlight the significant difference in control signal noise between RE+AL and the best agent trained by the previous Neuroflight training pipeline. Note that the RE+AL agents offer significantly smoother control, smoother even than the PID controller, while offering better tracking. This is further reflected in the lower power consumption of RE+AL agents, as shown in Table~\ref{MAE table}. }}
        \vskip -0.1in
        \label{fig:Freq}
    \end{figure}

\subsection{Test Performance}\label{EvalTest}

    \begin{figure}[t]
        \centering
        \includegraphics[width=.95\textwidth]{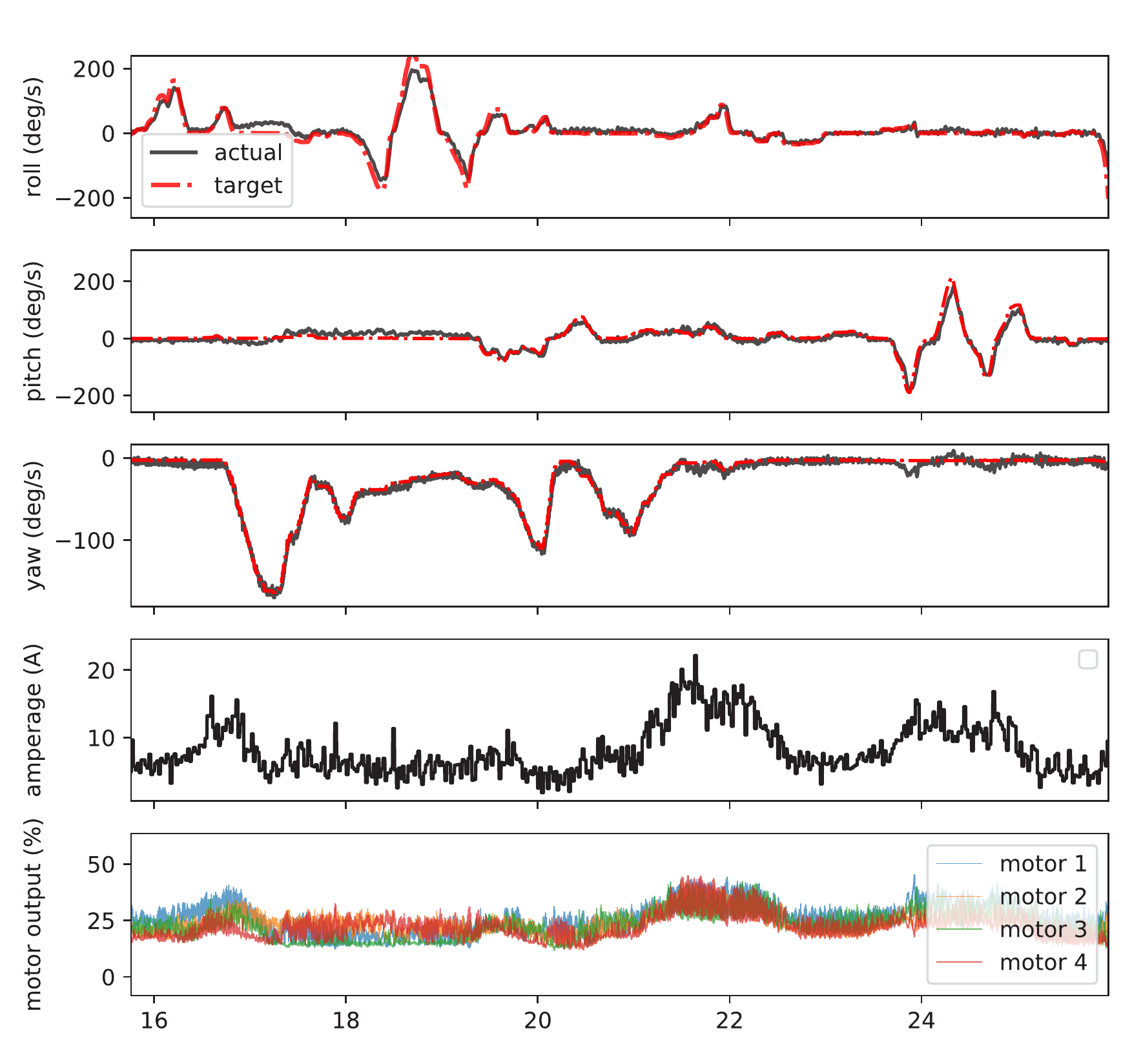}
        \caption{We show here a sample flight with a typical RE+AL agent -- While set-point tracking is generally successful, we observe that the motor control signals are noisier than in simulation but not enough to cause excessive spikes in the quadrotor's state response or power consumption.}
        \label{fig:ourflight}
    \end{figure}
    
    Upon completing training, the TensorFlow graphs representing each agent was compiled and deployed onto the drone, with which our pilot conducted a series of flight tests.
    Real-world flight tests focused on exploring the controllers' viable flight envelope, responsiveness, smoothness and power consumption.
    We compare our results with the best agent from the original Neuroflight framework.
    As Koch et al. noted in \cite{NFThesis}, they were only able to get one agent to successfully transfer a viable controller from simulation to reality while maintaining reasonable tracking performance.
    We were able to acquire a copy of this agent from the authors for comparison with agents trained by our method.
    Due to over-heating of the motors, however, our tests with the Neuroflight agent had to be limited in order to prevent hardware failures.
    
    We note from Table~\ref{MAE table} that our method produces agents that are comparable to the best agent previously developed by Neuroflight.
    As the Neuroflight training pipeline only produced one flight-worthy agent, we are unable to present a measure of variance for the Neuroflight agent.
    Crucially however, we are able to do this \emph{consistently}, with every agent trained by our method serving as a viable candidate for transfer to the real drone.
    Our method also produces agents that have significantly smoother control policies when transferred to the drone, as shown in Fig.~\ref{fig:Freq}.
    
    Fig.~\ref{fig:Freq} shows the average Fourier transform of the motor control signals across all our RE+AL agents compared against the best Neuroflight agent.
    Our agents have shown to produce smoother motor-control signals, with a peak output frequency of approximately 130Hz, compared to to Neuroflight's 330Hz.
    Note as well that the magnitude of the RE+AL agents' high frequency peak is significantly smaller than that of Neuroflight.
    In practice, this smoothness results in noticeably reduced heat generation in the motors, allowing us to fly for longer and without risking hardware failures due to motor burnout. A sample flight log is presented in Fig.~\ref{fig:ourflight}.
    The improved control output is further highlighted in the current draw during operation, with the motors of the RE+AL agents drawing, on average, {$5.86 \pm 3.10$ Amps during flight tests, while the Neuroflight agent drew an average of $22.87$ Amps on a similar hover test}.
    {This is also notably better than the average current draw of the PID controller.
    This implies that, not only are RE+AL agents capable of delivering smooth, stable and reliable control; but that they are also able to offer longer flight times, better even than a classical controller.
    To the best of our knowledge, our work is the the first to clearly demonstrate the improved practical utility of a neural-network controller over classical control on a problem where conventional wisdom might suggest that classical control techniques would get us closest to optimal control.}
    
\subsection{Analyzing The Reality Gap}\label{realityGap}
    
        \begin{figure}[b]
            \centering
            \vspace{-\baselineskip}
            \includegraphics[width=.75\textwidth]{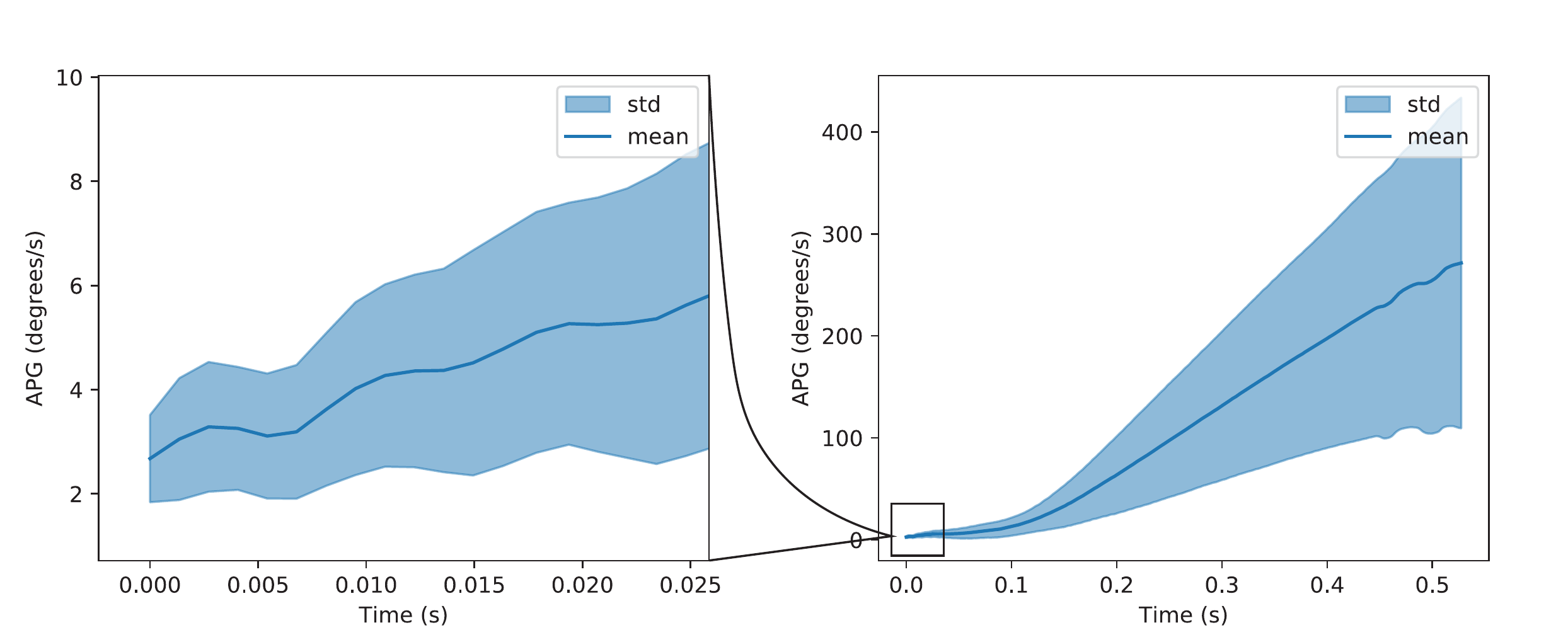}
            \vspace{-\baselineskip}
            \caption{
            \revision{We show here the Actuation Playback Gap between the rotational rate of the quadrotor in reality and in simulation when motor actuation from real flight is played back in simulation. While the drift is large over longer periods of time, it remains small for the time window of approximately 0.003s in which the controller responds. Measurements show are averaged over 160 samples.
            }
            }
            \label{fig:motor_gap}
            \vspace{-\baselineskip}
        \end{figure}
    
        \revision{
        A key factor limiting the efficacy of policy transfer of RL agents trained in simulation to real actuated platforms is the `reality gap', i.e. the discrepancy between real-life dynamics and their simulated approximations.
        If a simulator is a poor representation of reality, it stands to reason that the utility of behavior learned in simulation would be severely compromised upon transfer to a real platform.
        Conversely, with a perfect simulator, one should observe little to no difference in the performance of trained agents in simulation or reality.
        Our results, as highlighted in Table~\ref{MAE table} as well as Figs.~\ref{fig:Freq} and ~\ref{fig:ourflight}, demonstrate that RE+AL makes significant strides towards improving policy performance in transfer.
        However, the discrepancy in behavior observed when comparing Figs.~\ref{fig:Zoom}~and~\ref{fig:ourflight} indicates that the reality gap has not been completely overcome.
        }
        
        \revision{
        To better understand this gap and its impact on learning and policy transfer, we analyze the gap under two metrics: (i) \emph{actuation playback}: how the simulated model's response to logged actuation signals from real flight compares to the drone's actual behavior in flight, and (ii) \emph{set-point playback gap}: how trained controllers respond to the same control inputs in simulation and real flight.
        The first measure allows us to quantify and understand the limits of simulation fidelity, while the second allows us to quantify how different the system dynamics are for real and simulated systems from the perspective of the controller.
        }
        
        \revision{
        The actuation playback gap (APG) measures the discrepancy between how the physical drone responded to motor actuation in flight against how the simulated drone responds to the same actuation signals.
        To measure this, we sampled 160 0.5s trajectory segments from recorded flight logs.
        Given that actuation dynamics should be consistent regardless of controller, trajectory segments are sampled from all available flight logs with no consideration for which controller was used to generate it.
        For each segment, the simulator was initialized to the same state as the real drone at the start of the segment and the motor actuation signals were played back sequentially in simulation.
        The difference between the simulated and real states ($\phi_t^{sim}$ and $\phi_t^{real}$ respectively), with dynamics-constrained transition probabilities $P_{sim},\ P_{real}$, at time $t$ after initialization, is logged as:
        \begin{equation}
            APG_t = ||\phi_t^{real} - \phi_t^{sim}||_1  \ \ \text{for} \ \phi^{real}_{t} \sim P_{real}(\phi_t | \pi(s^{real}_{t-1})),\ \phi^{sim}_{t} \sim P_{sim}(\phi_t | \pi(s^{real}_{t-1})),\ s_0^{real} = s_0^{sim}
        \end{equation}
        }
        
        \revision{
        Fig.~\ref{fig:motor_gap} visualizes the trends in simulation dynamics drift over the 160 agents and shows how the discrepancy between simulated and real response grows as more time elapses from the point of initialization.
        While this may look catastrophically bad at first glance, it is also important to note the frequency at which the controllers operate.
        Our controllers operate at 730Hz, meaning that a control step is taken every 0.0014s (2 s.f.) and rely on information from the current and previous states.
        This implies that a trained controller only requires the simulated dynamics to be faithful to reality for a time window of approximately 0.0028s in order to be transferable.
        Over a time-horizon of 0.003s after initialization, the drift between simulation and reality is $3.05\pm 1.17$ deg/s, which is less than the tracking error threshold of the trained agents, allowing them to transfer reasonably well.
        To further verify this interpretation, we measured the set-point playback playback gap.
        }
        
\begin{figure}[h]
            \centering
            \includegraphics[width=\textwidth]{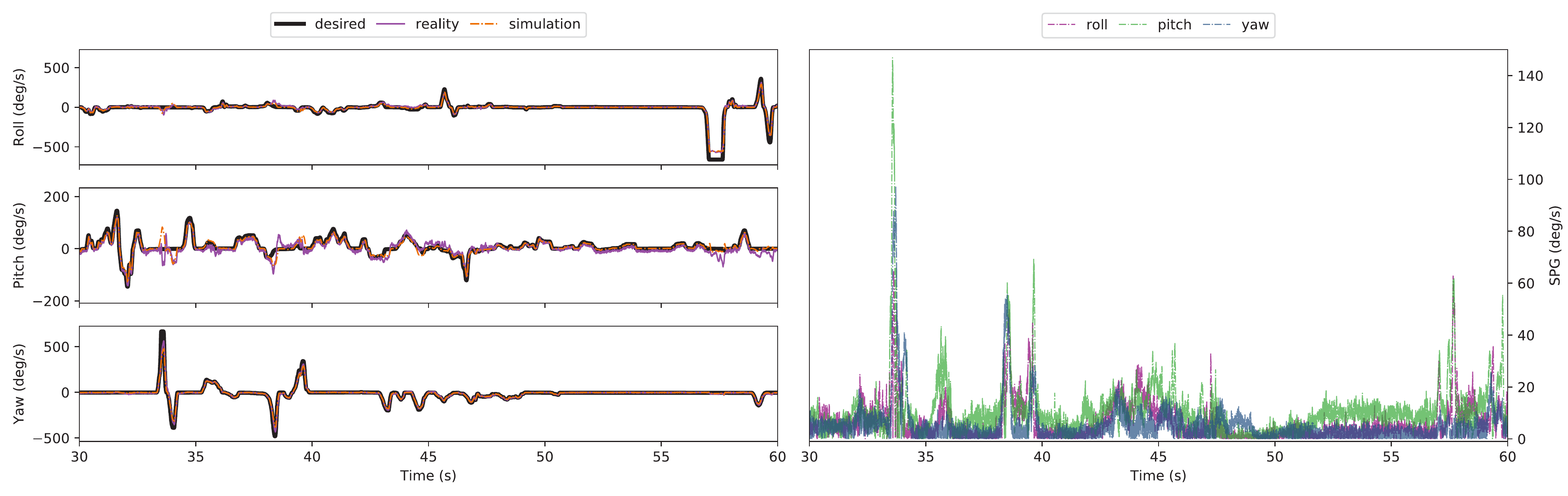}
            \caption{
            \revision{
            Sample Set-point Playback Gap (SPG) for a 30s flight segment where set-point signals received during flight are played back in simulation and the simulated and real flight trajectories are compared. We note that the difference is typically small, averaging 6.5deg/s, with spikes during more aggressive maneuvers. Averaged over multiple segments from multiple flights, we observe a set-point playback of 6.48 $\pm$ 6.06 deg/s.
            }
            }
            \label{fig:following_gap}
        \end{figure}

        \revision{
        The set-point playback gap (SPG) compares trajectories of a controller in real flight against the same controller with the same control inputs in simulated flight and is computed as:
        \begin{equation}
            SPG_t = ||\phi_t^{real} - \phi_t^{sim}||_1  \ \ \text{for} \ \phi^{real}_{t} \sim P_{real}(\phi_t | \pi(s^{real}_{t-1})),\ \phi^{sim}_{t} \sim P_{sim}(\phi_t | \pi(s^{sim}_{t-1})),\ s_0^{real} = s_0^{sim}
        \end{equation}
        This test measures the trajectory drift between a controller's response in flight and simulation.
        Should our interpretation hold, expected trajectory drift, $\mathbb{E}[SPG]$, in simulation would be bounded by the sum of the expected tracking error, $\mathbb{E}[MAE]$, and expected actuation playback gap, $\mathbb{E}[APG]$, i.e. $\mathbb{E}[SPG] \approx \mathbb{E}[MAE] + \mathbb{E}[APG]$, with its variance similarly bounded by $Var[SPG] \approx Var[MAE] + Var[APG]$ as the APG should be independent of specific controllers used. 
        This allows us to estimate that $SPG \approx 9.80 \pm 2.43$.
        Our experiments showed that the average trajectory drift was $6.48 \pm 6.06$, which falls within the expected range.
        Fig.~\ref{fig:following_gap}, shows the trajectory differences for a trained controller over a 30s SPG segment and illustrates how the simulated trajectory drift averages 6.50 deg/s, with spikes that occur during more aggressive maneuvers --- which is likely to include additional effects such as IMU measurement noise in addition to limitations in the controller's tracking capabilities.
        }

    \newpage
        
        
    \subsection{On the Extensibility of RE+AL Principles}\label{openAIGymEvals}
        \revision{
        We designed RE+AL primarily for the purpose of training more transferable policies.
        A key requirement of transferability is consistency of behavior.
        When training policies for the NF1 drone, the consistency of agents' performance is captured by the low variance in performance, which stands in stark contrast with training results produced by Koch et al.~\cite{NFThesis}.
        RE+AL's design consists of 3 main components: (i) multiplicative reward composition, (ii) state space tuning, and (iii) goal generation.
        Components (ii) and (iii) mainly ensure that agents have enough signal for the problem to be learnable and that training signals more closely relate to signals expected in real-world flight and mainly only apply to the specific problem of training quadrotor control.
        The main contributor to reducing performance variance however is the multiplicative reward composition, which we introduce as an alternative to the commonly used additive reward composition, and it is this component that we will demonstrate as having a broader utility.
        }
        
        \revision{
        To investigate if the benefits of multiplicative reward composition was limited in scope to our specific quadrotor control problem, or more generally applicable, we tested its effects on policies trained for 4 common open-source RL benchmarks~\cite{GYM}: HalfCheetah, Ant, Acrobot, and Pendulum.
        All 4 environments present with competing reward components are configured by default to use additive reward composition, with hand-tuned scalarization weights for each of their reward components.
        To test our method, We modified the reward computation functions for each environment to support multiplicative composition of normalized rewards and compared agents trained on vanilla additive rewards against the multiplicative rewards environment and vice-versa.
        We tested 3 commonly used contemporary RL algorithms: DDPG~\cite{DDPG}, PPO~\cite{PPO} and SAC~\cite{SAC}.
        While this is by no means an exhaustive collection of algorithms, it does allow us to compare algorithms from different schools of actor-critic RL: deterministic Q-learning, stochastic value-based policy gradient, and stochastic Q-learning respectively.
        As with training on the drone, we trained and tested multiple seeds for both additive and multiplicative reward composition on each of the benchmark environments.
        Results from our experiments are presented in Table~\ref{table:multiplicative}.
        }
    
        
        \begin{table*}[t]
            \vspace{-\baselineskip}
            \revision{
            \caption{
            \revision{
                Evaluations of Multiplicative Reward Composition on RL Benchmarks. Higher rewards are better while lower variance suggest more consistent learning performance. While multiplicative reward composition is not strictly superior to additive composition, it does yield more consistent performance and behavior.
            }
            }
            \label{table:multiplicative}
            \centering
            \setlength\tabcolsep{1.9pt}
            \begin{scriptsize}
            \begin{tabular}{c|||c|c||c|c||c|c}
                \hline
                Training & \multicolumn{6}{c}{Rewards $\uparrow$ $\pm\ \sigma \ \downarrow$ } \\ \cline{2-7}
                {Reward} & \multicolumn{2}{c||}{DDPG} & \multicolumn{2}{c||}{PPO}  & \multicolumn{2}{c}{SAC}  \\ \cline{2-7}
                Composition & Additive test & Multiplicative test & Additive test & Multiplicative test  & Additive test & Multiplicative test \\ \hline \hline
                \multicolumn{7}{c}{Half Cheetah} \\ \hline
                    Additive &
                    	$\pmb{2272.97} \pm 194.89$ & $655.98 \pm 39.93$ &
                    	$\pmb{1875.01} \pm 373.55$ & $602.07 \pm 59.95$ &
                    	$949.15 \pm 482.17$ & $233.98 \pm 216.72$ \\
                    Multiplicative &
                    	$1658.55 \pm \pmb{160.57}$ & $\pmb{695.93} \pm \pmb{23.07}$ &
                    	$1434.70 \pm \pmb{220.05}$ & $\pmb{643.88} \pm \pmb{43.79}$ &
                    	$\pmb{1259.80} \pm \pmb{185.69}$ & $\pmb{605.90} \pm \pmb{67.48}$ \\ \hline \hline
                \multicolumn{7}{c}{Ant} \\ \hline
                    Additive &
                    	$\pmb{1687.47} \pm 285.14$ & $766.09 \pm 28.59$ &
                    	$\pmb{361.86} \pm 606.21$ & $303.01 \pm 302.02$ &
                    	$\pmb{1299.70} \pm 226.23$ & $\pmb{742.88} \pm 38.93$ \\
                    Multiplicative &
                    	$1543.09 \pm \pmb{36.91}$ & $\pmb{784.87} \pm \pmb{2.78}$ &
                    	$198.20 \pm \pmb{30.95}$ & $\pmb{562.53} \pm \pmb{25.62}$ &
                    	$-138.38 \pm \pmb{39.44}$ & $196.58 \pm \pmb{6.43}$ \\ \hline \hline
                \multicolumn{7}{c}{Acrobot} \\ \hline
                    Additive &
                    	$462.95 \pm \pmb{64.65}$ & $35.64 \pm 32.65$ &
                    	$462.35 \pm 52.56$ & $22.31 \pm \pmb{7.76}$ &
                    	$512.60 \pm 47.28$ & $30.73 \pm \pmb{2.47}$ \\
                    Multiplicative &
                    	$\pmb{648.58} \pm 97.11$ & $\pmb{154.48} \pm \pmb{30.93}$ &
                    	$\pmb{514.05} \pm \pmb{44.24}$ & $\pmb{100.52} \pm 18.26$ &
                    	$\pmb{735.96} \pm \pmb{19.54}$ & $\pmb{180.25} \pm 5.25$ \\ \hline \hline
                \multicolumn{7}{c}{Pendulum} \\ \hline
                    Additive &
                    	$297.60 \pm 26.18$ & $68.98 \pm 15.62$ &
                    	$287.16 \pm 18.94$ & $\pmb{76.31} \pm 7.20$ &
                    	$\pmb{415.74} \pm \pmb{5.52}$ & $\pmb{120.18} \pm \pmb{1.20}$ \\
                    Multiplicative &
                    	$\pmb{349.17} \pm \pmb{12.19}$ & $\pmb{99.71} \pm \pmb{5.33}$ &
                    	$\pmb{312.34} \pm \pmb{1.53}$ & $75.68 \pm \pmb{0.76}$ &
                    	$404.67 \pm 6.85$ & $115.76 \pm 2.49$ \\ \hline
            \end{tabular}
            \end{scriptsize}
            }
            \vspace{-\baselineskip}
        \end{table*}
        
        \revision{
        As evidence for the consistency of multiplicative composition, we observe that, in 19 of the 24 comparisons, agents trained on multiplicatively composed rewards have lower variance than their additive composition counterparts.
        Furthermore, 13 of those 19 present with significant reduction in variance ($> 50\%$).
        Despite doing away with pre-existing tuning on reward components weights for additive scalarization, agents trained on a multiplicative composition demonstrate a clear ability to learn. 
        Unlike training with additive composition, agents trained with multiplicative reward composition typically better satisfied all the objectives defined in their respective environments, which is evidenced by the agents' often superior performance on nearly all tests on multiplicatively composed metrics.
        We also note that, in half of the tested cases, agents trained on multiplicative reward composition perform significantly better even on additive reward metrics.
        In cases where agents trained on multiplicative composition performed worse on the additive metric, we observed it was often due to agents not exploiting behaviors that satisfy some reward components at the cost of others (discussed further in Sections~\ref{hc_and_a}-\ref{acro}).
        This provides evidence that multiplicative composition, in addition to reducing variance, helps better ensure that agents train to better satisfy all reward components.
        Training on the multiplicative metric should not automatically be expected to yield `better' performance on the additive reward metric, but that it does allow for successful and often more consistent training, with room for tuning as necessary.
        While multiplicative reward composition is not strictly superior to additive composition, it does yield more consistent performance and behavior, and consistently better results with PPO (keeping with results observed in training agents for quadrotor control).
        }
    
        
        \subsubsection{Half Cheetah and Ant -- Environment-specific setup and discussion}\label{hc_and_a}
            \revision{
            Both Half Cheetah and Ant extend the HalfCheetahBulletEnv and AntBulletEnv environments provided with pybullet~\cite{pybullet}.
            Both are based on HalfCheetah-v2 and Ant-v2 respectively from the OpenAI Gym Benchmarks but use the open-source Bullet physics engine instead of the licensed Mujoco engine~\cite{todorov2012mujoco}.
            Their rewards comprise:
            \begin{enumerate}[(i)]
                \item \emph{Progress}: defined as the distance moved forwards during a step.
                \item \emph{Electricity usage}: defined as a cost in terms of the torque and velocities per joint.
                \item \emph{Joints at limit}: which counts the number of joints that are fully extended.
                \item \emph{Alive}: a binary flag indicating if the model has fallen.
            \end{enumerate}
            Training on the Half Cheetah environment allowed PPO and DDPG to learn more rewarding policies for rewards composed additively but this was found to be caused by agents exploiting behavior that resulted in increased progress at the cost of pushing more joints to their limits more often and consuming more resources.
            This is prevented in a multiplicative composition, which results in agents performing worse on the additive metric but better on the multiplicative one.
            SAC fails to train agents on the multiplicative reward composition on the Ant environment, which was found to be due to agents getting stuck in a minima early in training that they failed to break out of, though they were, in an ironic point in support of multiplicative composition, more consistent in learning behavior.
            We attempted to identify the cause of the minima but found that SAC agents failed to learn even with just a normalized Progress reward --- indicating that the core of the issue was not with the multiplicative composition but in how information about behavior was communicated to SAC agents by the Progress reward.
            This could likely be fixed with some tuning of the individual reward components, but was outside the scope of this study.
            Curiously, PPO fails on 2/3rd of trained agents by getting stuck on similarly `bad' behavior on additive reward composition, though with significantly higher variance.
            }
        
        \subsubsection{Pendulum -- Environment-specific setup and discussion}
            \revision{
            The Pendulum environment builds on the standard Pendulum-v0 OpenAI Gym environment and its rewards comprise:
            \begin{enumerate}[(i)]
                \item \emph{Stand}: how close the pendulum is to the upright position
                \item \emph{Velocity}: a penalty on the angular velocity of the pendulum
                \item \emph{Torque}: a penalty on the torque usage of the pendulum
            \end{enumerate}
            Both agents trained additively and multiplicatively successfully learn to balance the inverted pendulum.
            We did however observe that some agents would learn to ignore the velocity and torque objectives when trained on additive composition but always satisfied the objectives when trained with multiplicatively composed rewards.
            }
        
        \subsubsection{Acrobot -- Environment-specific setup and discussion}\label{acro}
            \revision{
            The Acrobot environment builds upon Acrobot-v1 from OpenAI Gym but is customized to support continuous control and dense reward signals.
            Its reward components comprise:
            \begin{enumerate}[(i)]
                \item \emph{Arm1 up}: the proximity the first arm has to the upright position
                \item \emph{Arm2 up}: the proximity the second arm has to the upright position
                \item \emph{Velocity1}: a penalty on the angular velocity of the first arm
                \item \emph{Velocity2}: a penalty on the angular velocity of the second arm
            \end{enumerate}
            Our version of Acrobot is the only environment where the trained agents were using exactly the same normalized rewards both multiplicatively and additively without tuning any weights. 
            All agents trained multiplicatively across all the algorithms have a higher average reward, whether tested with multiplicative or additive reward composition. 
            We observed that training on additively composed rewards results in agents getting stuck in local minima, while the agents trained multiplicatively all consistently learn to solve the problem by avoiding the minima.
            }

\subsection{Discussion}
    
    Based on our results, it is clear that RE+AL makes significant strides in improving the general viability of RL-based control on the task at hand.
    It is interesting to note from Table~\ref{MAE table} that both Neuroflight and RE+AL perform better in real flights (at least in tracking control set-points), as compared to tests in simulation -- a clear sign of the reality gap at work.
    What this implies is that albeit the GymFC simulator makes for a good representation of the dynamical environment, it is not without its limitations.
    This also further justifies our decision to practice early stopping during training.
    Though not reported here due to incomplete data, preliminary findings suggest that allowing RE+AL agents to train longer in simulation did not offer substantial benefits in transfer to the real quadrotor --- in fact, we observed that actuation would actually get noisier.
    This strongly suggests that further training would simply cause the agent to over-fit to the simulator, resulting in reduced transferability.
    
    Our RE+AL method is not without its drawbacks.
    A persistent issue observed both in training and testing is that the agents seem to be overly biased towards achieving good yaw tracking --- sometimes at the expense of good tracking along roll and pitch.
    This was also found to occasionally manifest as sudden perturbations in the quadrotor's angular roll or pitch velocities when a large change in angular yaw velocity was requested.
    While our pilot did not consider this to be a major issue and was able to quickly recover from any flight irregularities that resulted from this behavior, the fact remains that these occasional control glitches do exist in some agents.
    \revision{The general difficulty in reproducing other hardware-based pipelines makes it difficult to evaluate our method against related work but}
    as more generalized frameworks for training and testing RL-based control on hardware platforms become available, it would be interesting to evaluate RE+AL more thoroughly against alternative techniques such as those proposed by Hwangbo et al.~\cite{Hwangbo2017ControlOA} and Molchanov et al.~\cite{Sim2multi}.
    
    \revision{
    We also analyze the more fundamental benefits of the multiplicative reward composition introduced with RE+AL.
    Consistent with what we observed in training agents for quadrotor control, we note that multiplicative reward composition consistently helps reduce the variance in training RL agents across a variety of common RL benchmark tasks.
    While further study would be required before claims of general utility in RL training can be made, evidence suggests that multiplicative reward composition has a good compatibility with PPO training and can more broadly help simplify reward-engineering for performance consistency.
    }


\section{Conclusion}\label{conclusion}
    
    Our work introduces REinforcement-based transferable Agents through Learning (RE+AL), a framework for designing simulated training environments which encourage the training of smoother and more transferable controllers.
    RE+AL is built on a philosophy of improving the value of information provided to an RL agent during training while maintaining interpretability.
    Through the systematic design of descriptive yet simple reward signals, state space and environment-interaction paradigm, RE+AL facilitates the consistent training of RL agents with good control performance and reliable transferablity to real hardware.
    RE+AL is tested on the NF1 quadrotor platform, and we verified that 100\% of all agents trained with RE+AL were flight-worthy, offering good RC-control tracking and smooth actuation of the quadrotor's motors.
    RE+AL lays the groundwork for future work in developing controllers for more complex controls tasks, and also for incorporating additional learning techniques to further close the reality-gap.

\bibliographystyle{ACM-Reference-Format}
\bibliography{sample-base}

\end{document}